\documentclass[10pt,a4paper,twocolumn]{article}
\usepackage{bachelorproject}
\usepackage{graphicx}
\usepackage{hyperref}
\usepackage[table,xcdraw]{xcolor}

\usepackage{array}
\usepackage{mdwmath}
\usepackage{mdwtab}
\usepackage{eqparbox}
\usepackage{url}

\setcitestyle{square}

\title{\vspace{-0.5in}{\bfseries\scshape 
    Fine-grained 3D object recognition: an approach and experiments
    } \\ 
    \vspace{1ex} 
}

\author
{Junhyung Jo \quad Hamidreza Kasaei\\
\small University of Groningen\\
\small Nijenborgh 9, 9747 AG Groningen, The Netherlands \\
}
\date{\vspace{-5ex}}

\begin{document}

\twocolumn[
  \maketitle
  \begin{@twocolumnfalse}
    \begin{abstract}
Three-dimensional (3D) object recognition technology is being used as a core technology in advanced technologies such as autonomous driving of automobiles. There are two sets of approaches for 3D object recognition: (i) hand-crafted approaches like Global Orthographic Object Descriptor (GOOD), and (ii) deep learning-based approaches such as MobileNet and VGG. However, it is needed to know which of these approaches works better in an open-ended domain where the number of known categories increases over time, and the system should learn about new object categories using few training examples. In this paper, we first implemented an offline 3D object recognition system that takes an object view as input and generates category labels as output. In the offline stage, instance-based learning (IBL) is used to form a new category and we use K-fold cross-validation to evaluate the obtained object recognition performance. We then test the proposed approach in an online fashion by integrating the code into a simulated teacher test. As a result, we concluded that the approach using deep learning features is more suitable for open-ended fashion. Moreover, we observed that concatenating the hand-crafted and deep learning features increases the classification accuracy.
\end{abstract}

  \end{@twocolumnfalse}
]

\section{Introduction}\label{sec:introduction}
Object recognition is attracting attention as a core technology used in autonomous vehicles \cite{lee2020accuracy}, one of the industry sectors that are receiving attention in recent years, and is also used in various fields such as disease identification in bio-imaging, industrial inspection, and robot vision. Object recognition technology allow the robot to adapt to its new environment by improving its knowledge from the accumulation of experience and the conceptualization of new target categories. Through this, we can see that the importance of object recognition technology is gradually increasing, and accordingly, an attempt to improve the performance of object recognition technology is needed. Service robots used in the aforementioned industries are important for their ability to accurately recognize objects in complex environments and continue learning new categories. \textit{Open-ended object category learning} is described as the ability to learn new object categories sequentially without forgetting previously learned categories \cite{kasaei2021investigating}. Based on this theory, it seems like an attempt to create an interactive object recognition system is needed for learning 3D object categories in an open-ended fashion.

Before we talk about object recognition, it is needed to be compare the object recognition with \textit{object detection}, \textit{object representation} and \textit{perceptual memory}. 
\begin{figure}[!t]
    \centering
        \includegraphics[width=7.7cm]{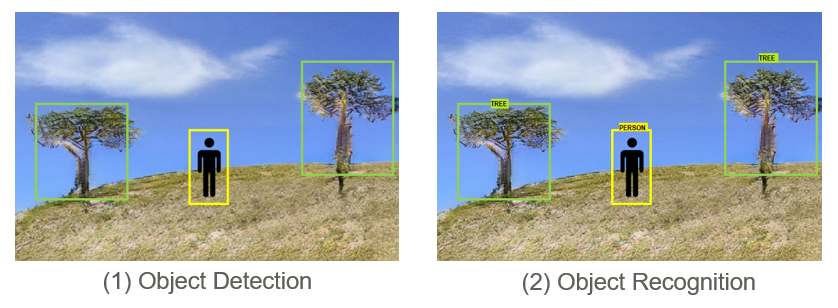}
    \caption{The example of (1) \textit{object detection} and (2) \textit{object recognition}.}
    \label{fig:obd}
    \vspace{-5mm}
\end{figure}

The 3D object recognition system we talked about earlier consists of these four software modules. Object detection is the process of finding an instance of an object in an image, whereas object recognition plays a role in recognizing it using the calculated values of the features of the detected object received from the object representation module. Objects are fully recognized by comparing them with the descriptions of existing known objects stored in perceptual memory.

The two main approaches to 3D object recognition that involve this process are including hand-crafted features, and deep-learning based methods. Approaches using hand-crafted features include Global Orthographic Object Descriptor (GOOD) \cite{kasaei2016good}, Ensemble of Shape Functions (ESF) \cite{wohlkinger2011ensemble}, and Viewpoint Feature Histogram (VFH) \cite{rusu2010fast}. These descriptors describe the shape and color of an object through its point cloud, RGB, and depth, and distinguish objects through them. Each descriptor uses a different frame of reference to compute a pose invariant description, which allows us to classify them. A detailed description of them will be provided later. The second main approach, the approach of deep learning models, is based on deep convolutional neural networks (CNNs). The deep learning approach shows effective performance when each category has a large amount of test data within fixed object categories. However, robots used in real-world situations need to learn and classify new categories in real time. Therefore, in this environment, robot will show limited performance. 
Looking at the two main 3D object recognition approaches, each has its pros and cons, and we do not know which approach performs better in an open-ended fashion, so it is needed to compare and analyze the two approaches to check their performance. Thus, we will take a look at how each approaches performs in recognizing and categorizing 3D objects in open-ended learning. The main research questions this paper seeks to answer can be:
\begin{flushleft}
    \textit{ Which learning progress of hand-crafted and deep transfer learning is the better for recognizing and categorizing of the 3D object in an open-ended fashion?}
\end{flushleft}
In this paper, we will create an interactive object recognition system that can learn 3D object categories in an open-ended fashion. This experience and accumulation of 3D object recognition and classification makes it possible to adapt to new environments by improving knowledge from conceptualization of new object categories. The task for this purpose is largely divided into two stages, and in each stage, the performance of the two approaches mentioned above, the approach using hand-crafted features and the approach based on deep transfer learning, is compared according to the given situation. Instance classification accuracy, class classification accuracy, and computation time, which are the results of the experiment, are used for performance comparison.

\begin{figure}[!t]
    \centering
        \includegraphics[width=\columnwidth]{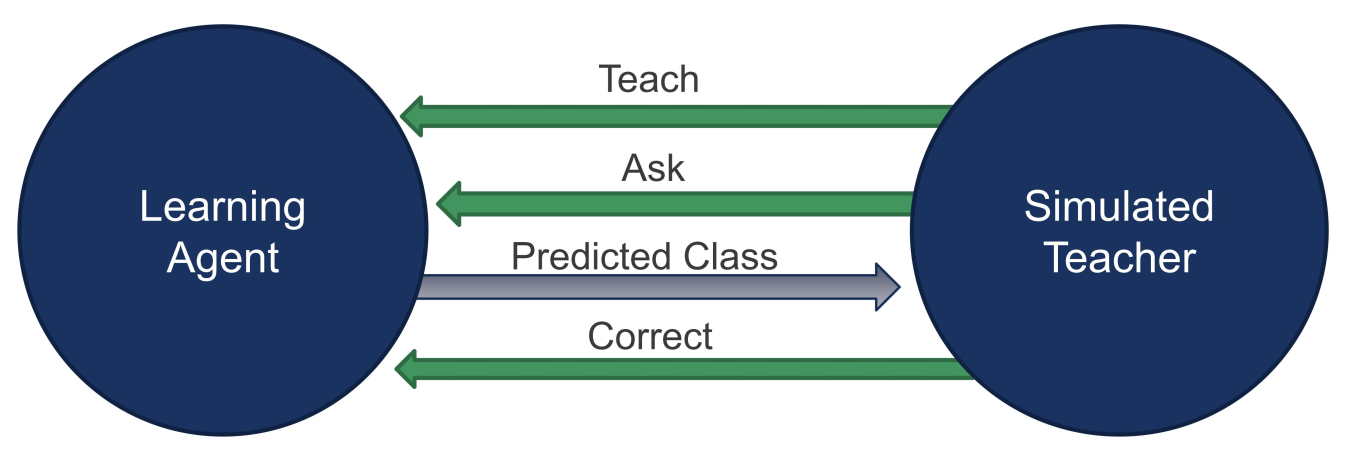}
    \caption{Simple architecture for the interaction between a simulated teacher and a learning agent \cite{of2020orthographicnet}.}
    \label{fig:teach}
    \vspace{-6mm}
\end{figure}

\begin{figure}[!t]
    \centering
        \includegraphics[width=7.7cm]{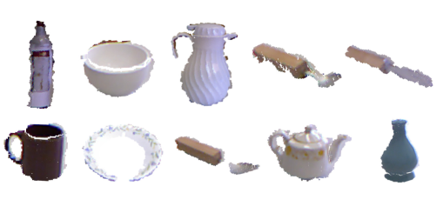}
    \caption{10 categorises of Restaurant RGB-D Object Dataset \citep{kasaei2015interactive}.}
    \label{fig:res}
    \vspace{-5mm}
\end{figure}

\begin{figure}[!b]
    \centering
        \includegraphics[width=7.7cm]{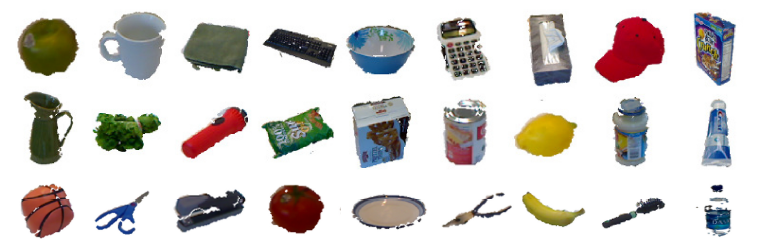}
    \caption{Some 3D point clouds of objects in Washington RGB-D Object Dataset \citep{lai2011large}}
    \label{fig:was}
\end{figure}

To give a brief overview of the stages that make up the research, in the first step, we begin by implementing an offline 3D object recognition system that takes an object view as input and generates category labels (e.g. apples, mugs, forks, etc.) as output. Then we will dedicate to testing our approach in an online fashion by integrating the code into a simulated teacher test. In the offline 3D object recognition system stage, instance-based learning (IBL) \cite{oliveira20163d} is used to form a new category and we use K-fold cross validation \cite{anguita2012k} to evaluate the obtained object recognition performance. In the open-ended stage which is the second step, it is determined that the evaluation method in the off-line is not suitable, and the teaching protocol designed for experimental evaluation in open-ended learning. In the off-line and open-ended stages, the Restaurant RGB-D Object Dataset (see Figure \ref{fig:res}) and Washington RGB-D Object Dataset (see Figure \ref{fig:was}) were used, respectively.

Once again, through this paper, we will mainly focus on detailing 3D object category learning and recognition, assuming that the object has already been segmented in the point cloud of the scene.

The remainder of this paper is organized in Sections 2 through 5 as follows. Section 2 describes works that related to this paper. The experimental details of this paper are described in section 3. The analysis of each approach according to the experimental results is covered in section 4. Finally, section 5 presents a conclusion and discusses future research.
\section{Related Work}\label{sec:work}
The performance of 3D object recognition systems has been continuously improved through several studies. These studies have also overcome the limitations of the two main approaches introduced above. In this section, we will look at studies related to approaches using hand-crafted features \cite{kasaei2021investigating}, approaches using deep learning features \cite{of2020orthographicnet}, and further approaches using concatenated features \cite{georgescu2019local}.
\subsection{Hand-Crafted Approach}
Most recent studies on approaches using hand-crafted features have dealt with only object shape information and ignoring the role of color information or vice versa. However, a recent study conducted by Kasaei et al explored the importance of not only shape information but also color constancy, color space and various similarity measures in open 3D object recognition \cite{kasaei2021investigating}. In this study, the performance of object recognition approaches was evaluated in different ways in three configurations, including color-only, shape-only, and color-and-shape combinations of objects. For this experiment, the color constancy information of the object was added to the GOOD object descriptor that did not contain the color information. 

\begin{figure}[!b]
    \centering
        \includegraphics[width=7.7cm]{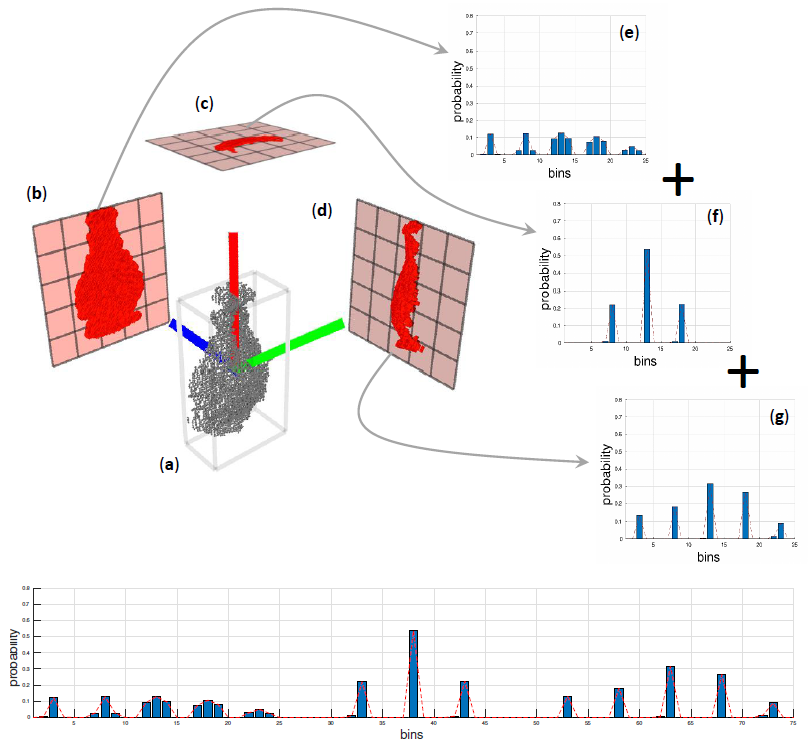}
    \caption{The process of GOOD object descriptors creating representations of `Vase' objects \citep{kasaei2021investigating}}.
    \label{fig:good}
\end{figure}

Figure \ref{fig:good} shows the steps to obtain an object representation from a vase object using GOOD with color information. The results of this study show that the combination of color and shape of an object leads to significant improvement in object recognition performance compared to shape-only and color-only approaches. This is because it is advantageous to distinguish objects with very similar geometric properties with different colors when color information is used together than when only shape information of the object is used. In addition, the approach combining color and shape information shows that robots can learn new object categories in real-world environments, even with a small number of training data.

\subsection{Deep Learning Approach}
\begin{figure}[!b]
    \centering
        \includegraphics[width=7.7cm]{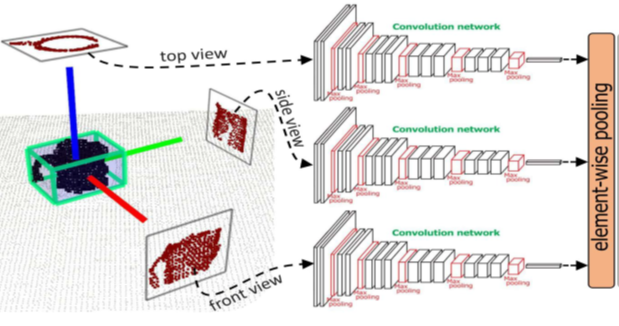}
    \caption{The process of the approach of deep transfer learning \citep{of2020orthographicnet}}.
    \label{fig:cnn}
\end{figure}
Another 3D object recognition approach, an approach using deep learning features, is based on CNNs. If the application model has a predefined set of fixed object categories and sufficient examples per category, an object recognition system built by training data through a deep CNN shows effective performance. However, CNNs have a limitation in that they are not structurally open-ended in a situation where learning and recognition are performed in real time, such as in the real world. In addition, if limited data is used for CNNs that require a large amount of data, it may lead to degradation of object recognition performance. Deep transfer learning can alleviate the limitations of CNNs by combining deep learning capabilities with online classifiers to handle open object category learning and recognition problems. Object recognition approaches using deep transfer learning can alleviate the limitations of approaches using CNNs by combining deep learning capabilities with online classifiers to handle open object category learning and recognition problems. \cite{of2020orthographicnet}. This approach, named OrthographicNet, computes a global object reference frame and three scale-invariant orthogonal projections of a given object, and merges the values computed via the max pooling function in the CNNs and uses it as a global feature (see Fig \ref{fig:cnn}). This approach via OrthographicNet has experimentally shown improvements over previous approaches via CNNs with respect to object recognition performance and scalability in open scenarios. This showed that the technique adopted by OrthographicNet showed the potential for performance improvement of approaches using deep learning features and could be more suitable than other approaches in real-time robotics applications.

\subsection{Concatenated Approach}
In previous related studies, it has been explained that the approach using hand-crafted features and the approach using deep learning features have their respective limitations. Attempts have been made to overcome these limitations, and the approach used by concatenating of the features of the previous two approaches is one of them. Georgescu et all showed that the approach using a feature concatenated with deep learning features and hand-crafted features showed more than 1\% higher accuracy than previous state-of-the-art approaches in facial expression recognition. \cite{georgescu2019local}.
\begin{figure}[h]
    \centering
        \includegraphics[width=7.7cm]{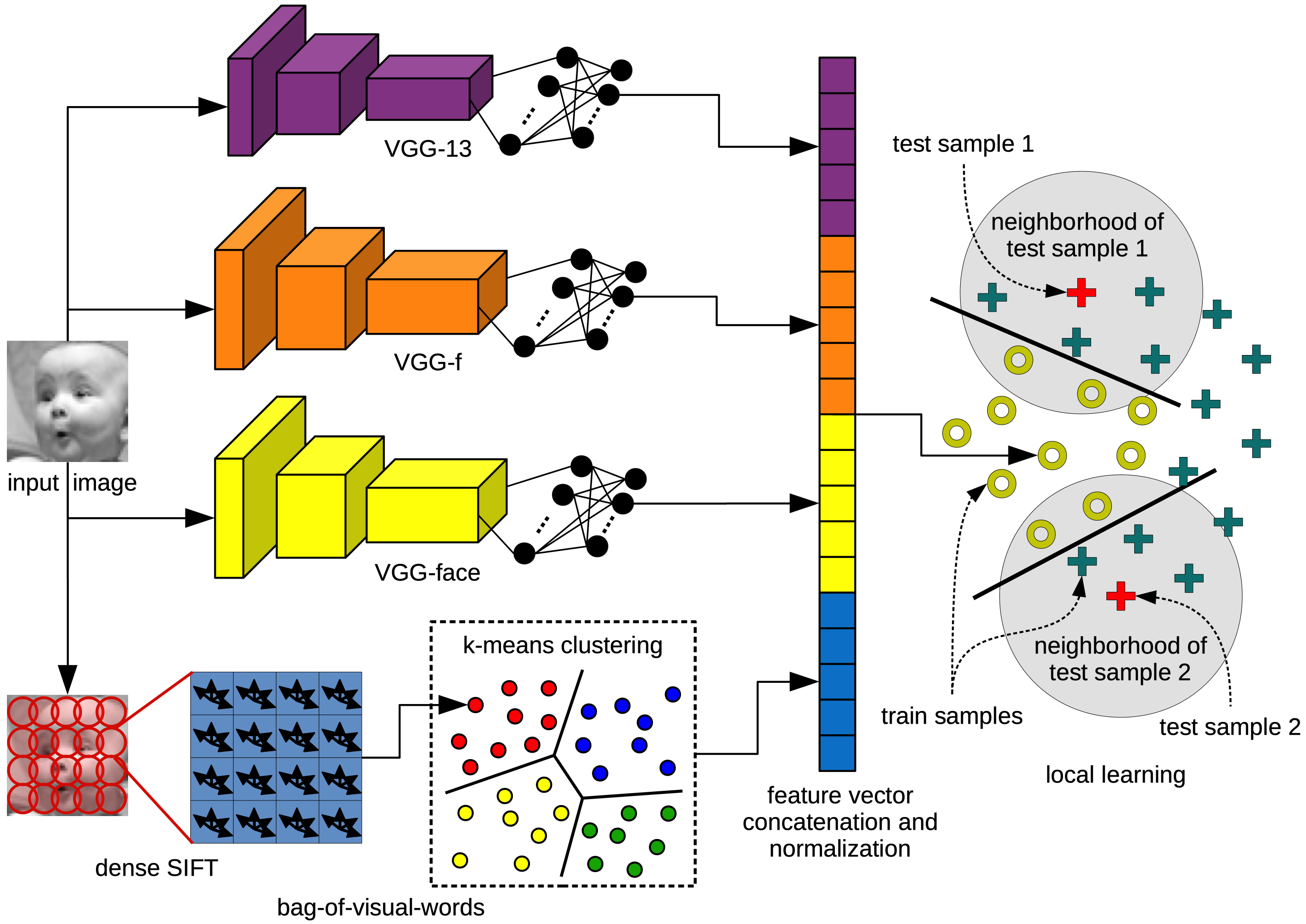}
    \caption{The process of the approach with concatenated feature on facial expression recognition \citep{georgescu2019local}}.
    \label{fig:conca}
\end{figure}

Deep learning features were acquired through CNN using Dense-Sparse-Dense (DSD) \cite{han2016dsd}, and a bag-of-visual-words (BOVW) model \cite{ionescu2013local} was used to obtain hand-crafted features. Additionally, Support Vector Machines (SVM) \cite{cortes1995support} based on the one-versus-all scheme was used for concatenating the corresponding feature vectors of the two features by combining the two models. The concatenated features thus obtained were normalized and used to recognize facial expressions (see Fig \ref{fig:conca}). The experimental results of this study show that the approach using combined features performs better than using one feature alone. This motivated this experiment by showing the possibility of obtaining the same results in 3D object recognition.
\section{Experiment}\label{sec:method}
The experiment of this research is divided into two main parts as follows.
\begin{enumerate}
    \item \textbf{Offline evaluation}: Implementing an offline 3D object recognition system to optimize basic 3D object recognition algorithms (Hand-crafted \& Deep learning), which takes an object view as input and produces as output the category label (e.g., apple, mug, fork, etc).
    \item \textbf{Online evaluation}: Testing the approaches in an online fashion by a simulated teacher test with obtained result from Part I.
\end{enumerate}
The K-fold cross validation used for the evaluation of the algorithm in each of these parts divides the data set into K equal-sized subsets, generating K folds randomly, each subset containing examples from all categories. The K-value was set to 10, and at each iteration, a single fold was used for testing and the remaining 9 folds were used for training data. This type of evaluation is useful not only for parameter tuning, but also for comparing the performance of different approaches and methods. We also used IBL as a way to form new categories of objects. The IBL approach can be viewed as a combination of object representation, similarity measures, and classification rules, which represent object categories as a set of object views of categories. In the case of similarity measurement, the similarity between objects can be calculated with different distance functions with a histogram normalized by the object descriptor (e.g. GOOD). In order to adopt dissimilar distance functions used in this process, referring to \cite{cha2007comprehensive}, the 14 distance functions were adopted and explored.

\subsection{Part I: Offline evaluation}
The purpose of the assignment in this part was to optimize at least four basic 3D object recognition algorithms and compare their results. Therefore, we prioritized finding two optimized algorithms for each approach, and this was done by tuning the various parameters applicable to each approach. The turning parameters that can be used for each approach are as follows. It should be noted that the deep network (e.g. \textit{mobileNet} \cite{howard2017mobilenets}) used as a parameter in the approach using deep learning features was trained on the imagenet dataset and we use it as a feature extraction tool. That means we don't train the network specifically for our application, we just use the output of one of the fully connected layers of the network as a function for the view of a given object. In addition, the parameter $K$ used in each approach below is the $K$ value used in the K-nearest neighbor (K-NN) algorithm \cite{peterson2009k}.
\begin{enumerate}
    \item \textbf{Hand-crafted object descriptor + IBL + K-NN}
    \begin{enumerate}
        \item Descriptors: [GOOD, ESF, VFH]
        \item Distance functions: [\textit{Euclidean, Manhattan, $x^{2}$, Pearson, Neyman, Canberra, KL divergence, symmetric KL divergence, Motyka, Cosine, Dice, Bhattacharyya, Gower, Sorensen}]
        \item $K \in [1, 3, 5, 7, 9]$
    \end{enumerate}
    \item \textbf{Deep transfer learning based object representation + IBL + K-NN}
    \begin{enumerate}
    \item Network architectures: [mobileNet, mobileNetV2 \cite{sandler2018mobilenetv2}, vgg16\_fc1, vgg16\_fc2, vgg19\_fc1, vgg19\_fc2 \cite{simonyan2014very}, xception \cite{chollet2017xception}, resnet50 \cite{he2016deep}, denseNet121, denseNet169, densenet201 \cite{huang2017densely}, nasnetLarge, nasnetMobile \cite{zoph2018learning}, inception \cite{szegedy2016rethinking}, inceptionResnet \cite{szegedy2017inception}]
    \item Element-wise pooling functions: [AVG, MAX, APP]
    \item $K \in [1, 3, 5, 7, 9]$
    \end{enumerate}
\end{enumerate}
The Restaurant RGB-D Object Dataset used here has a small number of classes with significant variability within the class. This is a suitable data set for conducting a wide range of experiments to tune the parameters of each approach, so it is suitable for conducting the experiments in this part.

The experiment was repeated by comparing the results obtained through K-fold cross-validation while tuning the parameters for each approach for the approach using the hand-crafted feature and the approach using the deep learning feature.

\subsection{Part II: Online evaluation}
The experiments in this part focus on finding the approach that shows better performance by comparing the best configuration of each approach obtained from offline evaluation in an open-ended fashion. Furthermore, the concatenation approach, which concatenates the features of the previous two approaches, was also tried and compared.
\begin{figure}[!b]
    \centering
        \includegraphics[width=7.7cm]{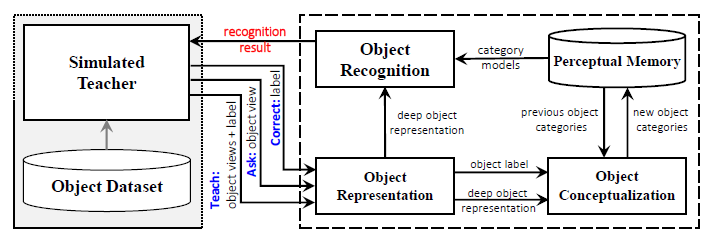}
    \caption{The architecture of the teaching protocol - Interaction between the simulated teacher (\textit{left}) and the learning agent (\textit{right}) \citep{of2020orthographicnet}}
    \label{fig:teach2}
    \vspace{-5mm}
\end{figure}

Unlike the previous offline evaluation, which does not follow the simultaneity of learning and recognition, by adopting a teaching protocol (see Fig \ref{fig:teach2}), we have created a suitable environment for evaluating open learning systems. The training protocol mimics and executes the interaction of the perception system with the surrounding environment over a long period of time in a single context scenario. The simulated teacher follows the training protocol and interacts with the learning agent using the three actions below.

\begin{enumerate}
    \item \textbf{Teach}: Introducing a new object category to the agent
    \item \textbf{Ask}: Ask the agent what is the category of a given object view
    \item \textbf{Correct}: Providing corrective feedback in case of misclassification
\end{enumerate}
Simulated\_teacher iteratively selects invisible objects from currently known categories and presents them to the learning agent via an \textbf{Ask} action for testing. If an object is represented as a training sample inside the learning agent by \textbf{Teach} or \textbf{Correct} instruction, it is stored in perceptual memory, otherwise it is passed to the object recognition module. Simulated\_teacher continuously estimates the agent's recognition performance using a sliding window of size $3n$ iterations. where $n$ is the number of categories already introduced. Finally, if $k$, the number of iterations since the new category was introduced, is less than $3n$, all results are used. If this performance exceeds a given classification threshold ($\tau = 0.67$, meaning that the accuracy is at least twice the error rate) \cite{kasaei2015interactive}, \cite{kasaei2018coping}, \cite{oliveira2015concurrent}, the teacher introduces a new object category by presenting three randomly selected object views. In this way, the learning agent is started with no knowledge and the instances are gradually made available by the training protocol.

In this part of the experiment, the Washington RGB-D Object Dataset with a larger amount of data than in offline evaluation was used. This dataset contains 300 objects in 51 categories, but only 200 objects in each category were used in this experiment.

The experiment proceeds by repeating K-fold cross validation for each best configuration of the two approaches taken from the previous offline evaluation and comparing the results. Since the experimental results may be affected by the system, it is repeated 10 times for each approach and compared with the average and standard deviation of the final result taken into account. In addition, the classification threshold $\tau$ mentioned above can also affect the experimental results, so we changed the classification threshold value to [0.7, 0.8, 0.9] and proceeded with the analysis.
\begin{figure*}[ht]
    \centering
        \includegraphics[width=15.4cm]{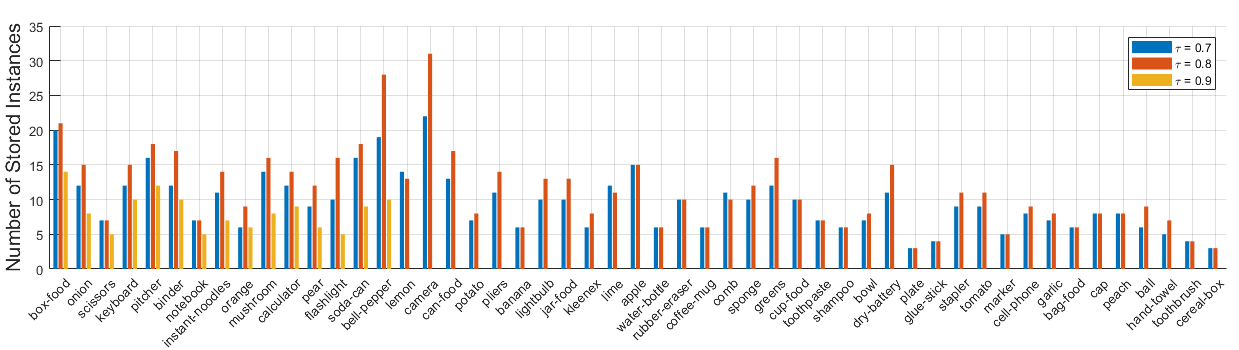}
        \includegraphics[width=15.4cm]{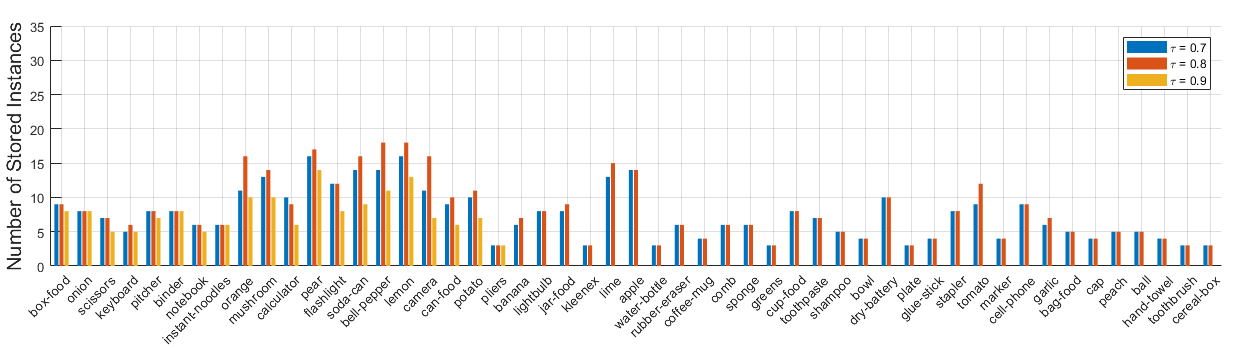}
    \caption{The result of the number of stored instance according to the classification threshold $\tau \in [0.7, 0.8, 0.9]$ on an approach with hand-crafted features (\textit{up}) and deep learning features (\textit{down}). In both approaches, it can be seen that a larger number of instances are classified when $\tau = 0.8$.}
    \label{fig:tau}
\end{figure*}

As a result of the analysis, when the $\tau$ is 0.8, considering that the value of the number of stored instances is higher (see Fig \ref{fig:tau}), so we set the $\tau$ to 0.8.
\section{Result \& Discussion}\label{sec:result}
In this section, the experimental results are explained according to the direction of each part of the experiment. In Part I, we aimed to find the best configuration of the approach using hand-crafted features and the approach using deep learning features. In Part II, we used them to compare the performance of the two approaches in an open-ended fashion. Furthermore, we tried an approach using concatenated features and compared it with the previous two approaches.
\subsection{Part I: Offline evaluation}
In offline evaluation, the parameters of each approach were tuned and the results were compared to find the best configuration of the two approaches using hand-crafted features and deep learning features.

\begin{figure}[ht]
    \centering
        \includegraphics[width=7.7cm]{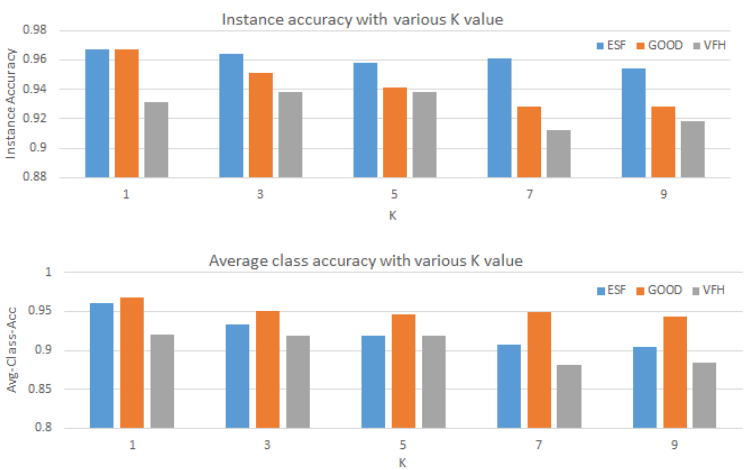}
    \caption{The results of the approach using hand-crafted features, the instance classification accuracy (\textit{up}) and average classification accuracy (\textit{down}).}
    \label{fig:hand}
    \vspace{-5mm}
\end{figure}

First, for the approach using hand-crafted features, the parameters of the selected object descriptor (e.g. GOOD, VFH, ESF) need to be tuned to achieve a good balance between recognition performance, memory usage, and processing speed. To this end, we tuned the distance function, which measures the similarity between the example object and the existing object, and the $K$ value used in K-NN as parameters. For the distance function, 14 functions introduced in Section 3.1 were searched and $[1, 3, 5, 7, 9]$ was used for the K value. For the GOOD object descriptor, the number of bins is a parameter, so the number of bins are searched from 10 to 100 at intervals of 10. The VFH object descriptor also has the normal estimation radius as a parameter, so it was searched from $1cm$ to $10cm$ at intervals of $1cm$. For ESF object descriptor, there are no special parameters. In the case of the distance function, when counting the number of times searched for each descriptor, $13\times3 = 42$ results came out, and after comparing the results, the \textit{Bhattacharyya} function was adopted as the common distance function. Using the \textit{Bhattacharyya} function as a common function, the result according to the parameter $K$ value of the three object descriptors is as follows (see Fig \ref{fig:hand}).

Comparing the above results, the object descriptor GOOD and ESF showed higher performance than other results with an instance classification accuracy of 96.74\% when the parameter $K$ was 1. However, the computation times of GOOD and ESF were 3.22s and 6.46s, respectively, and GOOD showed slightly better performance. Also, in average classification accuracy, it can be seen that GOOD shows an accuracy of 96.08\%  that shows better performance than other object descriptors.
\begin{figure}[ht]
    \centering
        \includegraphics[width=7.7cm]{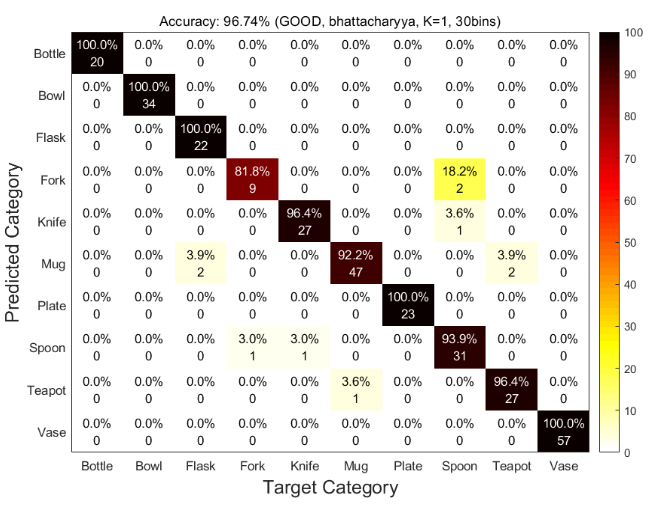}
    \caption{The confusion matrix of the best configuration of hand-crafted approach [GOOD, bhattacharyya, K=1, 30bins].}
    \label{fig:confu}
    \vspace{-5mm}
\end{figure}

In the Figure \ref{fig:confu}, which showed the best configuration in the approach using hand-crafted features, the parameter bin value of GOOD is 30. Setting a larger number of bins provides more detailed information about the point distribution, but increases computation time, memory usage, and sensitivity to noise, so 30 was chosen as a reasonable value. Therefore, [GOOD, bhattacharyya, K=1, 30 bins] was adopted as the best configuration in the approach using hand-crafted features.

Next, we will look at the experimental results of the approach using the deep learning feature. To find the best configuration for the approach using deep learning features, the 15 network architectures mentioned in section 3.1 were explored, and each network was explored for three element-wise pooling functions: AVG, MAX, and APP. Also, as with the hand-crafted approach, the $K$-values $[1, 3, 5, 7, 9]$ of K-NN were explored. Like the distance function in the previous result, the network architectures used in this part made $15\times3 = 45$ results. So we had to find the two best network architectures that performed better than the others, and as a result we finally adopted mobileNet and vgg16\_fc1.
\begin{figure}[ht]
    \centering
        \includegraphics[width=7.7cm]{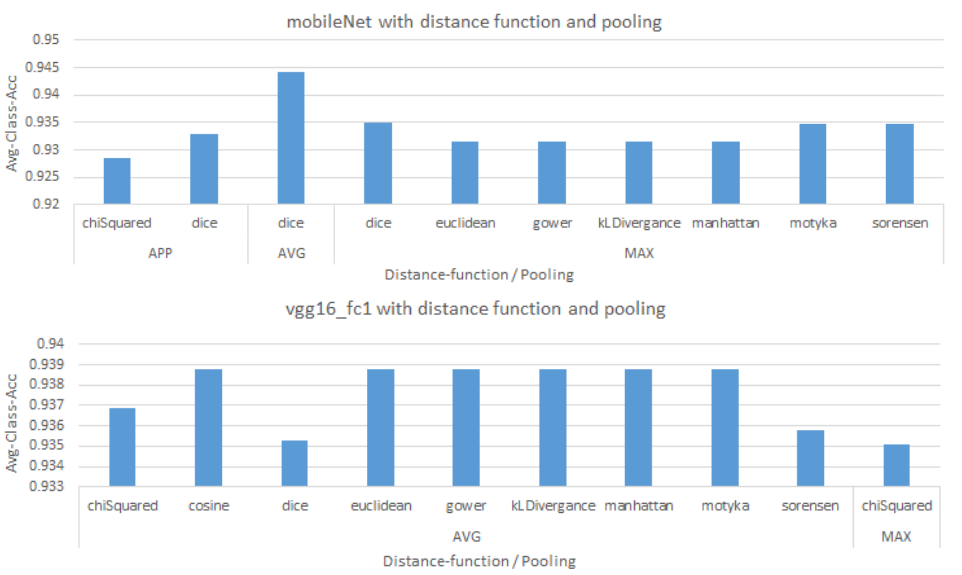}
    \caption{The results of the average classification accuracy of approach using deep learning features, mobileNet (\textit{up}) and vgg16\_fc1 (\textit{down}).}
    \label{fig:deep}
    \vspace{-5mm}
\end{figure}

As a performance result, the top 10 results are graphed for the adopted network architectures mobileNet and vgg16\_fc1, respectively, with the distance function and pooling function as parameters (see Fig \ref{fig:deep}). Looking at the graph, in the case of mobileNet, when the distance function \textit{dice} and the pooling function AVG are used, the average classification accuracy of 94.42\% is higher than that of other cases, and in the case of vgg16\_fc1, when 6 distance functions (\textit{cosin, euclidean, gower, kLDivergance, manhattan, motyka}) and AVG pooling function are used, 93.88\% accuracy was obtained. Therefore, [150, mobileNet, AVG, Dice] was adopted as the best configuration in the approach using deep learning features. Here, 150 in front of mobileNet means the orthographic image resolution used in the network.
\begin{figure}[!b]
    \centering
        \includegraphics[width=7.7cm]{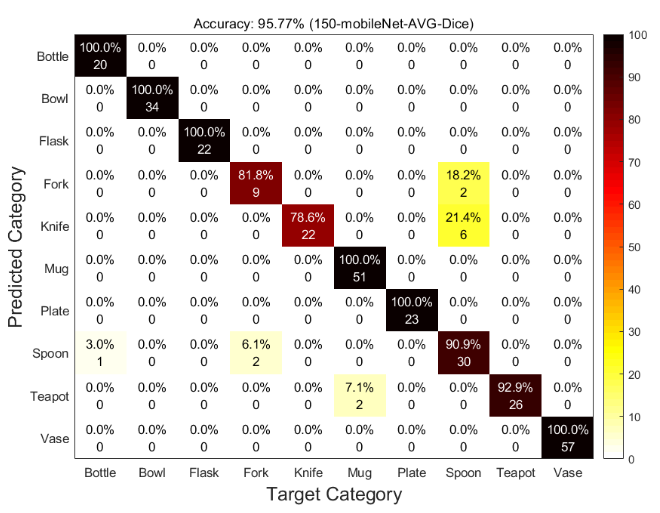}
    \caption{The confusion matrix of the best configuration of deep learning approach [150, mobileNet, AVG, Dice].}
    \label{fig:confu2}
\end{figure}
To summarize the experimental results of offline evaluation, [GOOD, bhattacharyya, K=1, 30 bins] was adopted as the best configuration in the approach using the hand-crafted feature, and in the approach using the deep learning feature, [150, mobileNet, AVG, Dice] has been adopted as the best configuration.Looking at the performance of the two approaches in this experimental result, in the case of average classification accuracy, the approaches using the hand-crafted feature and the deep learning feature were 96.08\% and 94.42\%, respectively, indicating that the hand-crafted approach performed better. This is because, in the offline evaluation, a hand-crafted approach designed specifically for encoding 3D objects shows better performance because the network was trained on the imageNet data set for natural images rather than orthogonal images.

\subsection{Part II: Online evaluation}
The average and standard deviation of the results obtained through 10 replicates experiments for comparing the best system configuration between hand-crafted and deep transfer learning approaches are represented in the Table \ref{tab:table}.
\begin{table}[!b]
\resizebox{7.7cm}{!}{%
\begin{tabular}{|c|r|r|r|r|}
\hline
\multicolumn{1}{|l|}{\cellcolor[HTML]{000000}{\color[HTML]{333333} }} & \multicolumn{2}{c|}{\textbf{Hand-Crafted}} & \multicolumn{2}{c|}{\textbf{Deep Learning}} \\ \hline
\textbf{Num} & \multicolumn{1}{c|}{\textbf{Ins.Acc}} & \multicolumn{1}{c|}{\textbf{Cls.Acc}} & \multicolumn{1}{c|}{\textbf{Ins.Acc}} & \multicolumn{1}{c|}{\textbf{Cls.Acc}} \\ \hline
1 & 0.7607 & 0.7667 & 0.8310 & 0.8400 \\ \hline
2 & 0.7570 & 0.7797 & 0.8356 & 0.8465 \\ \hline
3 & 0.7648 & 0.7777 & 0.8317 & 0.8430 \\ \hline
4 & 0.7569 & 0.7731 & 0.8392 & 0.8600 \\ \hline
5 & 0.7493 & 0.7750 & 0.8274 & 0.8395 \\ \hline
6 & 0.7619 & 0.7720 & 0.8430 & 0.8634 \\ \hline
7 & 0.7703 & 0.7876 & 0.8634 & 0.8823 \\ \hline
8 & 0.7493 & 0.7584 & 0.8385 & 0.8637 \\ \hline
9 & 0.7814 & 0.8038 & 0.8604 & 0.8786 \\ \hline
10 & 0.7521 & 0.7608 & 0.8218 & 0.8293 \\ \hline
\rowcolor[HTML]{ECF4FF}
Avg & 0.7603 & 0.7755 & 0.8391 & 0.8547 \\ \hline
Std & 0.0099 & 0.0132 & 0.0135 & 0.0177 \\ \hline
\end{tabular}%
}
\caption{The results of 10 experiments with two approaches (hand-crafted approach \& deep learning approach) with Ins.Acc (Instance classification accuracy) and Cls.Acc (Class classification accuracy).}
\label{tab:table}
\end{table}

In online evaluation, the approach using deep learning features showed 7.88\% and 7.92\% higher performance in instance classification accuracy and class classification accuracy, respectively, than the approach using hand-crafted features.
\begin{figure*}[ht]
    \centering
        \includegraphics[width=\linewidth, trim={4cm 0cm 4cm 0cm}, clip=true]{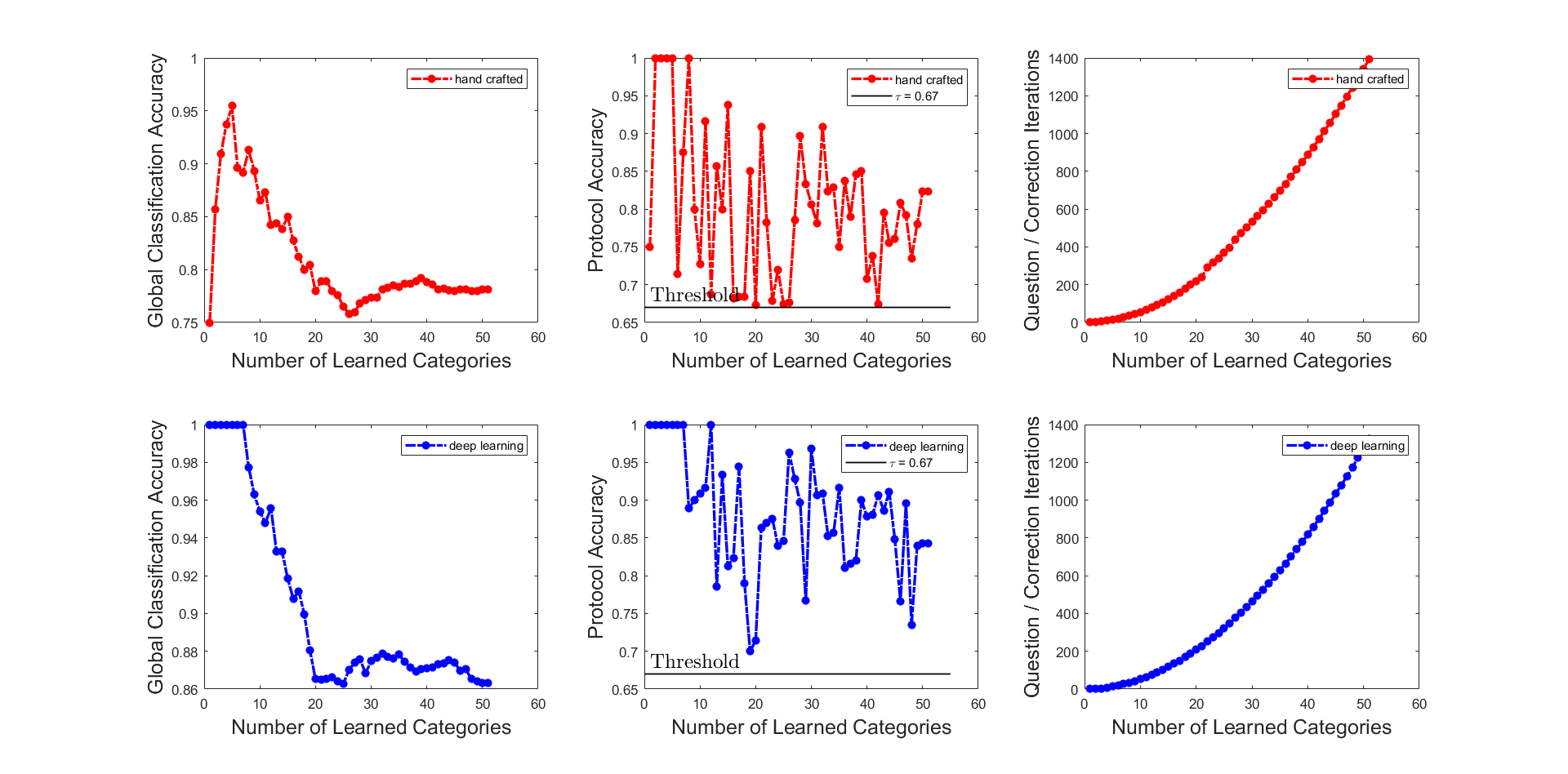}
    \caption{Global classification accuracy, protocol accuracy, and question/correction iterations according to the number of learned categories of approaches using hand-crafted features (\textit{up}) and approaches using deep learning features (\textit{down}).}
    \label{fig:comb}
\end{figure*}

In addition, if you look at the Figure \ref{fig:comb}, the approach using the deep learning feature shows at least 86\% in global classification accuracy, while the method using the hand-crafted feature shows an accuracy of less than 80\% in the section where the number of learned categories is between 20 and 50. Even in protocol accuracy, approaches using hand-crafted features often approach the threshold, whereas approaches using deep learning features safely exceed the threshold.

The following is a comparison between the previous two approaches and the approach using concatenated features.
\begin{table}[!t]
\centering
\resizebox{4.4cm}{!}{%
\begin{tabular}{l|cr}
\cellcolor[HTML]{FFFFFF}{\color[HTML]{333333} } & \multicolumn{2}{c}{\textbf{Concatenated}} \\ \cline{2-3} 
\multirow{-2}{*}{\cellcolor[HTML]{FFFFFF}{\color[HTML]{333333} }} & \textbf{Ins.Acc} & \multicolumn{1}{c}{\textbf{Cls.Acc}} \\ \hline
\rowcolor[HTML]{ECF4FF} 
\multicolumn{1}{c|}{\cellcolor[HTML]{ECF4FF}{\color[HTML]{000000} Avg.}} & \multicolumn{1}{r}{\cellcolor[HTML]{ECF4FF}0.9547} & 0.9397 \\ \hline
\rowcolor[HTML]{ECF4FF} 
\multicolumn{1}{c|}{\cellcolor[HTML]{ECF4FF}{\color[HTML]{000000} Std.}} & \multicolumn{1}{r}{\cellcolor[HTML]{ECF4FF}0.0100} & 0.0152
\end{tabular}%
}
\caption{The results of average and standard deviation of 10 experiments with an approach using concatenated features (hand-crafted + deep learning).}
\vspace{-5mm}
\label{tab:table2}
\end{table}

In the Table \ref{tab:table2}, we can see that the approach using concatenated features shows about 10\% higher accuracy in instance classification accuracy and class classification accuracy than the previous two main approaches. In other words, when the 3D object recognition system uses concatenated features, it shows better performance than when hand-crafted features or deep learning features are used alone. In this experimental result, we confirmed that when we concatenate hand-crafted features and deep learning features, the performance varies depending on the ratio between the two features, and when the ratio of deep learning features is 80\% to 90\%, it shows better performance than when it is not. The ratio of features used by both approaches affects the distance function, which, as explained earlier, is concerned with the similarity between the example object and the training object. For example, if there is an example object \textit{a} and a training object \textit{b}, the similarity between the two objects can be expressed as a distance function $D$.
\begin{center}
    $D(a, b) = (1 - w) \times D_h(a, b) + w \times D_d(a, b)$
\end{center}
For this experiment, $D_h$ is the distance function of the approach using hand-crafted features, so \textit{bhattacharyya} is the distance function. So $D_d$ becomes \textit{dice} distance function. The weight $w$ indicates the importance of the object representation of the two approaches, so in the concatenated approach, the ratio between the two approaches indicates the effect of the approach on the similarity between the two objects. Thus, by using concatenated features, hand-crafted features encode the curvature and shape information of an object, whereas deep learning features encode depth image-related features such as edges and intensity, which can be missed by using each approach alone. For these reasons, we got better performance.
\section{Conclusion}\label{sec:conclusion}
In this paper, we compared the performance of the two main 3D object recognition approaches, the approach using hand-crafted features and the approach using deep learning features, and attempted to find an approach that shows better performance for open-ended environments. Furthermore, an approach that concatenates the features of the two approaches was also tried and compared with the existing two approaches. This was an attempt to find an answer to the research question introduced above. In this process, we found the best configuration of the two main approaches through offline evaluation, and we checked and compared the performance of the two approaches in an open-ended environment through online evaluation using them. As a result, it was confirmed that the approach using the deep learning feature showed better performance in the open-ended environment than the approach using the hand-crafted feature. In addition, the approach using a feature concatenated with hand-crafted features and deep learning features showed a 10\% improvement in instance classification accuracy and class classification accuracy compared to the previous approach using deep learning features, showing new possibilities. Therefore, we came to the conclusion that the 3D object recognition system shows better performance in an open-ended environment when using the concatenated feature than when using the hand-crafted feature and the deep learning feature alone.

\subsection{Further research}
The approach using concatenated features used in this paper used features from the two approaches adopted through offline evaluation. However, this is not a conclusion reached by examining all configurations of the approach using the hand-crafted feature and the approach using the deep learning feature, so there remains the possibility of improving the performance through other configurations. For example, if the vgg16\_fc1 network architecture is concatenated with the ESF object descriptor, better performance may be obtained, and furthermore, a network or object descriptor not introduced here may be used. Combinations such as vgg-13, vgg-f, and bag-of-visual-words used for facial expression recognition introduced earlier can show good performance in 3D object recognition as well. As many new deep learning networks and approaches using hand-crafted features are currently being studied, the conclusions covered in this paper can be sufficiently changed with further research. Therefore, we should study with this in mind.

\bibliographystyle{IEEEtran}
\bibliography{literature}

\clearpage

\end{document}